\documentclass[lettersize,journal]{IEEEtran}
\usepackage{amsmath,amsfonts}
\usepackage{algorithmic}
\usepackage{algorithm}
\usepackage{array}
\usepackage[caption=false,font=normalsize,labelfont=sf,textfont=sf]{subfig}
\usepackage{textcomp}
\usepackage{stfloats}
\usepackage{url}
\usepackage{verbatim}
\usepackage{graphicx}
\usepackage{cite}
\usepackage[dvipsnames]{xcolor}
\usepackage{comment}
\usepackage{booktabs}

\begin{document}

\title{GlueNN: gluing patchwise analytic solutions with neural networks}
\author{Doyoung Kim, Donghee Lee, Hye-Sung Lee, Jiheon Lee, and Jaeok Yi
\thanks{Doyoung Kim, Donghee Lee, Hye-Sung Lee, Jiheon Lee, and Jaeok Yi are with the Department of Physics, Korea Advanced Institute of Science and Technology, Daejeon 34141, Korea (email: ehdud923@kaist.ac.kr; dhlee641@kaist.ac.kr; hyesung.lee@kaist.ac.kr; anffl0101@kaist.ac.kr; wodhr1541@kaist.ac.kr)}
\thanks{ All authors contributed equally. Author names are listed in alphabetical order.}}

\maketitle

\begin{abstract}
In the analysis of complex physical systems, the objective often extends beyond merely computing a numerical solution to capturing the precise crossover between different regimes and extracting parameters containing meaningful information. However, standard numerical solvers and conventional deep learning approaches, such as Physics-Informed Neural Networks (PINNs), typically operate as black boxes that output solution fields without disentangling the solution into its interpretable constituent parts. In this work, we propose GlueNN, a physics-informed learning framework that decomposes the global solution into interpretable, patchwise analytic components. Rather than approximating the solution directly, GlueNN promotes the integration constants of local asymptotic expansions to learnable, scale-dependent coefficient functions. By constraining these coefficients with the differential equation, the network effectively performs regime transition, smoothly interpolating between asymptotic limits without requiring ad hoc boundary matching. We demonstrate that this coefficient-centric approach reproduces accurate global solutions in various examples and thus directly extracts physical information that is not explicitly available through standard numerical integration.
\end{abstract}

\begin{IEEEkeywords}
GlueNN, differential equation, patchwise analytic solutions
\end{IEEEkeywords}

\section{Introduction}

Differential equations arise across physics, engineering, and applied mathematics, providing a compact mathematical language for expressing fundamental laws and describing the evolution of complex systems. Solving these equations is therefore central to both theoretical understanding and quantitative prediction. In practice, however, obtaining analytic solutions can be challenging when the dynamics are nonlinear and complicated.

In such situations, deep learning has emerged as a complementary tool that can approximate solutions directly from data and/or physical constraints. Among deep learning based approaches, physics-informed neural networks (PINNs)~\cite{RAISSI2019686} have become a useful tool. PINNs embed the governing equations directly into the loss function to recover a specific solution for a given input configuration. Various studies ~\cite{JAGTAP2020113028,osti_2282003,jin2022asymptoticpreservingneuralnetworksmultiscale, Bertaglia2022Asymptotics,HU2023107183,Pratama2023EXPLORING,Amirhossein2023Theory, wang2024aspinnasymptoticstrategysolving, Huang2024MultiScale, zhu2025deepasymptoticexpansionmethod, LIU2025113669, shen2025matchedasymptoticexpansionsbasedtransferable, sun2025pvdonetmultiscaleneuraloperator, Pradanya2025Parameter, MICHOSKI2020193, Huang2023NeuralStagger, hu2024betterneuralpdesolvers, verma2025cosmologyinformedneuralnetworksinfer, Song:2020agw, Jeong:2025omu} have advanced this approach.

While deep learning models can approximate solutions with high accuracy, their outputs are often difficult to interpret, which limits their practical utility in scientific applications. In many problems, the objective extends beyond obtaining solutions at discrete points; instead, one seeks insight into the underlying dynamics, such as dominant behaviors or conserved quantities. Accordingly, methods that emphasize interpretable representations—rather than treating the model as a purely black-box function—are particularly valuable when the goal is to understand the governing mechanisms, not merely to obtain accurate numerical predictions.

To address this issue, we build on the observation that many differential equations contain multiple competing terms whose relative importance varies across the domain. In practice, different regions are governed by distinct leading terms, so that local analytic solutions can be obtained by solving simplified equations in each regime. These regime-wise solutions are typically easier to interpret conceptually, since the dominant term makes transparent which effects control the dynamics and how the solution scales with parameters. The remaining challenge is then to assemble the local solutions into a single global solution that satisfies the given initial or boundary conditions. This step is often nontrivial because, near the interface between regimes, the approximation schemes break down. Consequently, matching based solely on continuity of the solution and a finite number of derivatives at a chosen boundary point can introduce systematic errors. 

In this study, we address this challenge by proposing the \textbf{GlueNN} framework. GlueNN achieves a smooth connection between asymptotic solutions by promoting the coefficients in the general asymptotic forms to learnable parameters. During training, the network learns to regulate these coefficient functions so that they approach finite constants within the regime where each asymptotic expansion is valid, while smoothly suppressing the corresponding contributions in regimes where they should not apply. This mechanism naturally bridges the intermediate region, yielding physically meaningful predictions across the entire domain without an ad hoc matching scheme. By using the pattern recognition capability of deep learning, GlueNN infers the nontrivial transition dynamics between different regimes and enforces a smooth and consistent interpolation between asymptotic limits.

The contributions of this study are summarized as follows.
\begin{enumerate}
\item We propose a structured neural architecture that directly extracts patchwise analytic components, enforcing interpretability by design rather than through post-hoc procedures.
\item We introduce a composite training objective that combines (i) initial-condition fitting, (ii) enforcement of the full differential-equation residual over the domain, and (iii) an out-of-patch suppression penalty, thereby avoiding ad hoc boundary matching and stabilizing cross-regime behavior.
\item We validate the framework on representative problems ranging from chemical kinetics to quantum tunneling, demonstrating its broad applicability across diverse equations and problem structures, including settings with multiple linearly independent solution and multiple regime crossings, through accurate global predictions and improved performance relative to conventional matching procedures.
\end{enumerate}

The remainder of this paper is organized as follows. In Section~2, we describe the problem statement and the details of the GlueNN framework. Section~3 introduces the example problems and presents the results obtained using this method. Finally, Section~4 provides a discussion.

\section{Related Works}

\subsection{Augmented PINN (APINN)}

APINN ~\cite{HU2023107183} introduces a flexible subdomain-wise decomposition by combining multiple subnetworks through a gating mechanism. In particular, APINN uses gating networks to impose soft domain decomposition, and the final prediction is formed as their weighted superposition. This design enables the model to adaptively allocate representational capacity across the domain and to mitigate training difficulties associated with strongly heterogeneous solutions. However, the decomposition learned by the gate is primarily functional rather than analytic: the subnetworks are generic function approximators, and the resulting representation is a smooth mixture whose components are not constrained to scientifically meaningful asymptotic solutions. Consequently, while APINN is effective for adaptive domain decomposition, it does not explicitly encode asymptotic analytic structure, nor does it guarantee that each subnetwork captures a regime with a direct physical interpretation.

 In contrast, we begin from analytic asymptotic forms that are valid in distinct regimes and promote their integration constants to learnable functions. Our training procedure is designed so that these coefficient functions approach finite constants within their corresponding validity ranges, while being suppressed outside those ranges. This yields an interpretable representation in which each component preserves a clear meaning.

\subsection{Asymptotic Properties in Deep Learning}

A number of studies have explored asymptotic approaches in neural networks for function extrapolation~\cite{Antonov2020AsymptoticsControl, routray2025enforcingasymptoticbehaviordnns}, as well as deep learning methods for multiscale partial differential equations, often in the presence of a small perturbation parameter~\cite{jin2022asymptoticpreservingneuralnetworksmultiscale, Bertaglia2022Asymptotics,Pratama2023EXPLORING, Amirhossein2023Theory, wang2024aspinnasymptoticstrategysolving, Huang2024MultiScale, zhu2025deepasymptoticexpansionmethod, LIU2025113669, shen2025matchedasymptoticexpansionsbasedtransferable, sun2025pvdonetmultiscaleneuraloperator, Pradanya2025Parameter}. These approaches typically aim to improve generalization and numerical robustness by enforcing prescribed far-field behavior, by designing architectures/losses that remain accurate across disparate scales, or by incorporating asymptotic/multiscale priors that are parameterized through expansions. The regime separation and the learning objectives in such methods are often tied to the existence (or knowledge) of a perturbation parameter and its associated asymptotic hierarchy, and the learned components do not necessarily map to explicit integration constants of analytic asymptotic solutions.

In contrast, our framework targets asymptotic structure that emerges along the domain independently of any perturbation parameter and incorporates this structure directly through the integration constants of the asymptotic solutions.

\begin{figure*}[t]
    \centering
    \includegraphics[height=95mm]{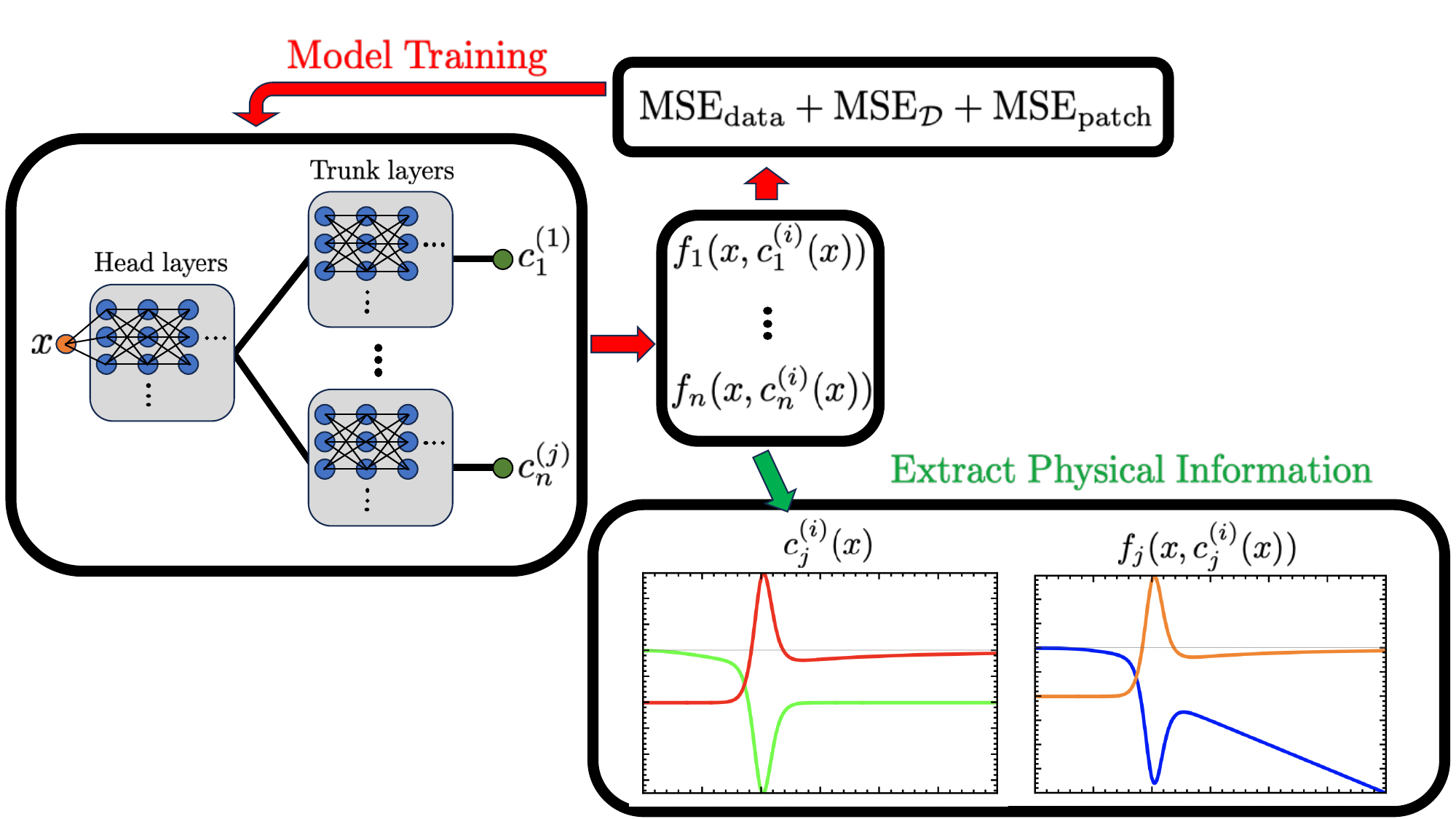}
    \caption{Schematic of the GlueNN framework. The input coordinate \(x\) is processed by a shared head network and multiple trunk branches, each producing a coefficient function \(\{c_j^{(i)}(x)\}\). These coefficients parameterize the patchwise analytic forms \(f_j\bigl(x;\{c_j^{(i)}(x)\}\bigr)\), whose sum defines the global ansatz. Training minimizes a composite loss function, \(\mathrm{MSE}_{\mathrm{data}} + \mathrm{MSE}_{\mathcal{D}}\), enforcing agreement with data and the differential-equation residual over the full domain. The learned coefficients and functions provide an interpretable view of how the solution transitions between asymptotic regimes. }
    \label{fig:Schem}
\end{figure*}

\section{Methods}

Consider a complicated differential equation governed by a differential operator $\mathcal D$ such that 
\begin{equation}
\mathcal D [f(x)]=0,
\end{equation}
where $f(x)$ denotes the unknown solution.
In general, deriving an exact analytic expression of $f(x)$ is difficult. However, in many problems, the governing differential equation exhibits distinct asymptotic behaviors depending on the values of the independent variable $x$. In each patch of the domain, certain terms can become negligible relative to others, allowing the complex full equation to be significantly simplified. Consequently, approximate analytic solutions may be constructed within distinct subdomains. 

As an example, consider a case in which the equation admits two characteristic asymptotic forms defined in separate patches, $\{p_1: x\ll a\}$ and $\{p_2: x\gg a\}$. In these regions, the dynamics is effectively described by:
\begin{equation} 
\mathcal D_1[f_1(x,{c_1^{(i)}})]=0 \quad \text{and} \quad \mathcal D_2[f_2(x,{c_2^{(i)}})]=0,
\end{equation} 
where $f_1(x,{c_1^{(i)}})$ and $f_2(x,{c_2^{(i)}})$ represent the general analytic solutions in their respective patches, and $c_1^{(i)}$ and $c_2^{(i)}$ denote the sets of undetermined coefficients to be fixed by boundary or matching conditions. 

Such an analysis can offer powerful qualitative insight into phenomena, since the analytic forms $f_1$ and $f_2$ often contain physical information, or the undetermined coefficients may represent physically conserved quantities. However, this approach faces significant challenges in the transition regions between patches. While it is standard practice to connect asymptotic analytic solutions by imposing matching conditions to ensure continuity (and, if necessary, $n$th-order differentiability), the outcome often depends arbitrarily on the choice of the matching point. Improving the match typically requires empirical, time-consuming tuning and may demand additional information.

GlueNN is designed to address this problem by utilizing deep neural networks (see Fig.~\ref{fig:Schem}), while preserving the underlying philosophy of asymptotic analysis. Unlike standard PINNs, which directly approximate the full solution $f(x)$, GlueNN promotes the coefficients $c_1^{(i)}$ and $c_2^{(i)}$ to $x$-dependent functions. These functions then become the trainable outputs of the neural network. The objective of this framework is to find coefficient functions, i.e., $c_j^{(i)}(x)$, such that the global ansatz
\begin{equation}
y(x) = \sum_j f_j\!\left(x,{c_j^{(i)}(x)}\right),
\end{equation}
satisfies the full equation to within a prescribed tolerance $\epsilon$,
\begin{equation}
    |\mathcal{D}[y(x)]| < \epsilon,
\end{equation}
over the domain of interest. One naturally expects that each $f_j(x, c_{j}^{(i)}(x))$ vanishes outside the regime in which it provides the appropriate asymptotic description, and that the learned coefficients $c_j^{(i)}(x)$ approach constant values deep inside each patch.

Training GlueNN requires a loss function $\ell$ that quantifies how well the network’s prediction matches the target solution. The loss function combines Mean Squared Error (MSE) terms corresponding to (i) solution mismatch at sample points, (ii) deviation from the differential equation, and (iii) an out-of-patch suppression penalty, and can be written as
\begin{equation}
\ell=\text{MSE}_{\mathrm{data}}+ \text{MSE}_{\mathcal{D}} + \text{MSE}_{\mathrm{patch}}.
\label{eq:loss}
\end{equation}
Each MSE term is evaluated on a distinct set of points,
$\{x_\alpha\}$, $\{x_\beta\}$, and $\{x_\gamma\}$:
\begin{equation}
\text{MSE}_{\mathrm{data}}
= \frac{\lambda_a}{N_a}\sum_{i,\,j,\,  x \in\{ x_\alpha\}}^{} 
\left| f_j \!\left(x, c_j^{(i)}(x)\right) - h(x) \right|^2,
\label{eq:msedata}
\end{equation}
\begin{equation}
\text{MSE}_{\mathcal{D}}
=\frac{\lambda_b}{N_b} \sum_{i,\,j,\, x\in \{x_\beta\}}^{} 
\left| \frac{\mathcal{D}\!\left[f_j\!\left(x, c_j^{(i)}(x)\right)\right]}{\mathcal{N}\!\left[f_j\!\left(x, c_j^{(i)}(x)\right)\right]} \right|^2,
\label{eq:msed}
\end{equation}
\begin{equation}
\text{MSE}_{\mathrm{patch}}
= \frac{\lambda_c}{N_c}\sum_{i,\,x\in\{x_\gamma\}}^{} 
\left| f_j\!\left(x, c_j^{(i)}(x)\right) \right|^2,
\label{eq:msepatch}
\end{equation}

where $N_a$, $N_b$, and $N_c$ denote the number of sample points in each set, and $h$ is the target solution, which is available only within the patch where the initial condition is given. We introduce $\mathcal{N}$ as a normalization function to control the scale of the differential-equation loss across the domain.  Accordingly, to incorporate the information required to solve the differential equation, the points $\{x_\alpha\}$ used to evaluate $\mathrm{MSE}_{\mathrm{data}}$ are sampled from that initial condition patch.

In contrast, the points to evaluate $\text{MSE}_{\mathrm{patch}}$ are drawn from outside the region where each patch ($j$) specific solution is valid, thereby suppressing its contribution in regimes where it should be inactive. Equation~\eqref{eq:msepatch} can be straightforwardly extended to multiple sample sets, each serving to suppress different asymptotic solutions outside its domain of validity. The hyperparameters $\lambda_a$, $\lambda_b$, and $\lambda_c$ control the relative weights of the MSE terms and must be chosen to balance the contributions of the three components in the total loss.

We employ a head–trunk architecture (see Fig.~\ref{fig:Schem}), in which separate trunk branches are assigned to each parameter, allowing them to learn distinct behaviors. Our implementation builds on the open-source PINNs-Torch package~\cite{bafghi2023pinnstorch}, which we have modified to support the GlueNN architecture and the coefficient-based parameterization used in this work.

\section{Experiments}

In this section, we apply GlueNN to three representative examples in which the solution exhibits distinct asymptotic regimes. These experiments show that GlueNN provides accurate global solutions while yielding coefficient functions that encode the underlying physics across regimes.

\subsection{Chemical Reaction}

\begin{table}[tbp]
\centering
\caption{Sampling Intervals and Hyperparameters for Chemical Reaction Experiment}
\label{tab:chemical_params}
\setlength{\tabcolsep}{6pt} 
\begin{tabular}{ >{\centering\arraybackslash}p{0.14\columnwidth} | >{\centering\arraybackslash}p{0.50\columnwidth} | >{\centering\arraybackslash}p{0.20\columnwidth} }
\toprule 
\textbf{Set} & \textbf{Interval} & \textbf{Points ($N$)} \\
\midrule \midrule 
$\{x_\alpha\}$ & $[1.0, 1.9]$ & $28$ \\
$\{x_\beta\}$ & $[7.8, 31]$ & $60$ \\
\bottomrule 
\multicolumn{3}{p{0.95\columnwidth}}{\scriptsize Note: The loss weights were set to $\lambda_a=1600$ and $\lambda_b=700$. All the sample sets are collected logarithmically in the given interval.}
\end{tabular}
\end{table}

In our first example, we investigate a reversible chemical reaction where the abundances of species evolve through generic processes of the form
\begin{equation}
    A + B  \Leftrightarrow C + D.
\end{equation}
For a given setup, the corresponding differential equation can be derived from the underlying reactions governing the situation. 

Here, we focus on a chemical reaction in a hot thermal plasma that cools monotonically with time. As the temperature decreases, the reactions that maintain chemical equilibrium become inefficient, and the abundance of the target species freezes out to an asymptotic value. The yield $Y\equiv n_A/s$ of species $A$ satisfies a specific form of a Riccati equation~\cite{Kolb:1990vq}
\begin{equation}
   \mathcal{D}[Y(x)] = \frac{dY}{dx} +\eta x^{-2}(Y^2-Y_{\mathrm{eq}}^2) =0,
   \label{eq:freeze_diff}
\end{equation}
where $n_A$ is the number density of $A$ and $s$ is the entropy density of the thermal bath. Here $x\equiv m/T$, with $m$ the mass of $A$ and $T$ the plasma temperature, and $\eta x^{-2}$ parameterizes the normalized interaction rate as a function of $x$. Since $T$ decreases monotonically in time, we can use $x$ as a time variable.

At early times, corresponding to small $x$, interactions are efficient and enforce chemical equilibrium, so $Y \simeq Y_{\mathrm{eq}}$, where\footnote{The numerical prefactor follows from the equilibrium distribution of a non-relativistic Bose or Fermi gas~\cite{Kolb:1990vq}.}
\begin{equation}
    Y_{\mathrm{eq}}(x) = 0.145\, x^{3/2} e^{-x}.
\end{equation}
However, as the plasma cools, the interaction rate becomes too small to maintain chemical equilibrium, so the abundance of species $A$ departs from $Y_{\mathrm{eq}}$ and saturates to a constant value.

For training, we used a head network with two hidden layers and output layers of widths 100, together with two trunk networks, each with one hidden layer of width 100, followed by an output layer of width 1. The output is parameterized as
\begin{equation}
    y(x) = e^{c_1^{(1)}(x)} x^{3/2} e^{-x} + e^{c_2^{(1)}(x)},
    \label{eq:sol_chem}
\end{equation}
where $c_{1}^{(1)}$ and $c_{2}^{(1)}$ are the outputs of the two trunk networks. 
We exponentiate the outputs to enforce positivity and to make it easier for the network to learn coefficients spanning several orders of magnitude. The MSE losses are constructed as
\begin{align}
\mathrm{MSE}_{\mathrm{data}} &= \frac{\lambda_a}{N_a}\sum_{x\in\{x_\alpha\}}|y(x) - h(x)|^2,\\
\mathrm{MSE}_{\mathcal{D}} &= \frac{\lambda_b}{N_b}\sum_{x\in\{x_\beta\}} \left|\frac{\mathcal{D}[y(x)]}{y(x)}\right|^2, 
\end{align}
where $\mathcal{D}$ is given in Eq.~\eqref{eq:freeze_diff} with $\eta =10^{4}$.  

\begin{figure}
\centering
\subfloat[\label{fig:freez_res_a}]{\includegraphics[width=0.99\linewidth]{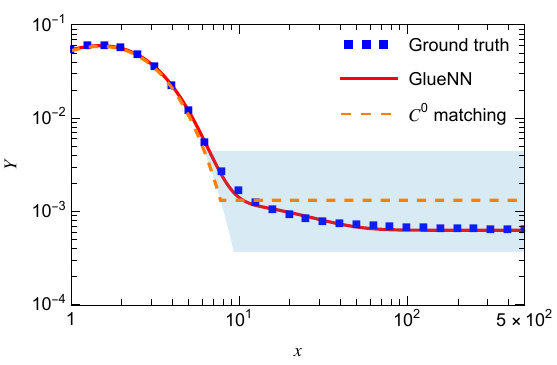}}\hfill
\subfloat[\label{fig:freez_res_b}]{\includegraphics[width=0.99\linewidth]{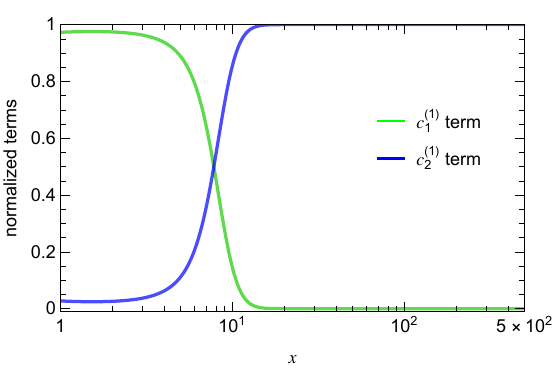}}\hfill
\caption{Training results for the chemical reaction example. (a): The ground truth value of the field amplitude, the prediction from the GlueNN, and the curve obtained from the $C^0$ matching procedure. The shaded band indicates the variation obtained by shifting the matching point by $\pm 20\%$. GlueNN closely reproduces the true solution.   (b): The normalized coefficients of the asymptotic solutions, which approach constant values toward the end of the domain. Here, each term in Eq.~\eqref{eq:sol_chem} is normalized by $y(x)$ itself. }
\label{fig:freez_res}
\end{figure}

We present the hyperparameter for the experiment in Table~\ref{tab:chemical_params}, and the training results in Fig.~\ref{fig:freez_res}. The GlueNN prediction closely matches the true solution, even though no training data are provided for $x> 1.9$. It also reproduces the expected asymptotic behavior: the $c_1^{(1)}$ term dominates at small $x$, while the $c_2^{(1)}$ term dominates at large $x$. The $c_2^{(1)}$ term can therefore be used directly to extract the final yield of the target species. 

By contrast, if one manually matches two analytic solutions at the interface between patches using $C^0$ matching\footnote{We define $C^n$ matching scheme as enforcing continuity of the function value and its derivatives up to order $n$ at a chosen matching point between two distinct functions.}, the matched value is fixed by \(Y_{\mathrm{eq}}\) evaluated at the chosen matching point and then propagated into the adjacent patch. Because \(Y_{\mathrm{eq}}\) depends exponentially on \(x\), the resulting manual matching is exponentially sensitive to the matching location and hence numerically unstable. As shown in Fig.~\ref{fig:freez_res_b}, a $\pm20\%$ shift in the matching point from $x=7.8$ leads to a substantial change in the final yield. Consequently, manual matching cannot be used to reliably determine the target species' yield, whereas GlueNN offers a robust alternative.

\subsection{Inflationary Production of a Vector Particle}

In particle physics and cosmology, the origin of dark matter is one of the central open problems. Among many candidates, one possibility is a massive vector boson associated with a hidden $U(1)$ gauge symmetry, often referred to as a dark photon \cite{Holdom:1985ag,Fabbrichesi:2020wbt,Caputo:2021eaa}. Such a particle naturally arises in extension of the Standard Model with additional gauge sectors.

In the minimal setup, the vector boson can be produced from the quantum fluctuation during the primordial inflation of the universe, and the evolution of the field amplitude of the vector boson is described by the following equation \cite{Graham:2015rva}.

\begin{table}[tbp]
\centering
\caption{Sampling Intervals and Hyperparameters for Inflationary Production Experiment}
\label{tab:inflation_params}
\setlength{\tabcolsep}{6pt} 
\begin{tabular}{ >{\centering\arraybackslash}p{0.14\columnwidth} | >{\centering\arraybackslash}p{0.50\columnwidth} | >{\centering\arraybackslash}p{0.20\columnwidth} }
\toprule 
\textbf{Set} & \textbf{Interval} & \textbf{Points ($N$)} \\
\midrule \midrule 
$\{a_\alpha\}$ & $[0.1, 1.0]$ & $10$ \\
$\{a_\beta\}$ & $[1.0, 124]$ & $650$ \\
$\{a_\gamma\}$ & $[23, 500]$ & $382$ \\
\bottomrule 
\multicolumn{3}{p{0.95\columnwidth}}{\scriptsize Note: The loss weights were set to $\lambda_a=800$, $\lambda_b=0.50$, and $\lambda_c=0.0032$. The set $\{a_\alpha\}$ is collected linearly, while $\{a_\beta\}$ and $\{a_\gamma\}$ are collected logarithmically.}
\end{tabular}
\end{table}

\begin{multline}
  \mathcal{D} \left[X_L(a)\right]=\frac{d^2X_L}{da^2} +\frac{2}{a}\frac{dX_L}{da}+\bigg\{\frac{k^2}{H^2 a^4} + \frac{m^2}{H^2a^2} \\ -\frac{k^2}{k^2+m^2a^2}\left( \frac{2}{a^2}-\frac{3m^2}{k^2+m^2a^2}\right)\bigg\}X_L =0,\label{eq:vectorEq} 
\end{multline}
where $H$ is the Hubble parameter, which is constant during inflation, $m$ the mass, $k$ the comoving momentum of the vector boson, and $a$ is the scale factor that parameterizes the expansion of the universe. As $a$ evolves, the hierarchy among the terms inside the braces can change. In this case, one can approximate the expression in the braces of Eq.~\eqref{eq:vectorEq} in two distinct patches:
\begin{equation}
    \begin{cases}
        \dfrac{k^2}{H^2a^4}-\dfrac{2}{a^2}\quad &\mathrm{for}\quad \mathcal{S}_1,\\  \dfrac{k^2}{m^2a^2} \quad &\mathrm{for}\quad \mathcal{S}_2, 
    \end{cases}
\end{equation}
where we assume that $m^2\ll kH$, and two patches are defined as  $\mathcal{S}_1: a\ll k/m$, $\mathcal{S}_2:  a\gg k/m$.

Then the solution at each patch is
\begin{equation}
X_L(a) = \begin{cases}
 c_1^{(1)} H^{(1)}_{3/2}\left(\dfrac{k}{aH}\right) +c_1^{(2)} H^{(2)}_{3/2}\left(\dfrac{k}{aH}\right) \quad &\mathrm{for} \quad \mathcal{S}_1,\\ 
 c_2^{(1)} J_{-1/2}\left(\dfrac{k}{ma} \right) + c_2^{(2)} J_{1/2}\left(\dfrac{k}{ma} \right)\quad &\mathrm{for} \quad\mathcal{S}_2,
    \end{cases}
    \label{eq:X_L_asymt}
\end{equation}
where $H^{(1)}_{3/2}$ and $H^{(2)}_{3/2}$ are Hankel's functions of the first and second kind, while $J_{-1/2}$ and $J_{1/2}$ are Bessel functions of the first kind.  In the training example, we set $c_1^{(2)}=0$, since it is the given initial condition of the vacuum~\cite{Bunch:1978yq}.

\begin{figure}
\centering
\subfloat[\label{fig:grav_res_a}]{\includegraphics[width=0.99\linewidth]{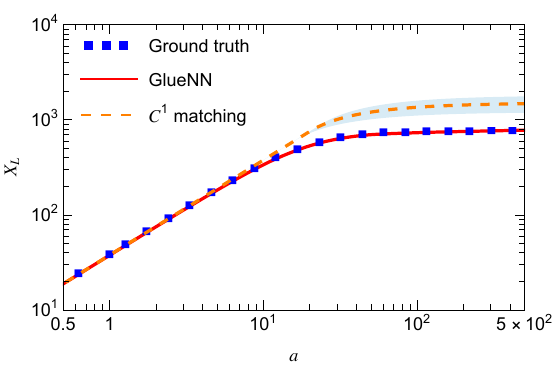}}\hfill
\subfloat[\label{fig:grav_res_b}]{\includegraphics[width=0.99\linewidth]{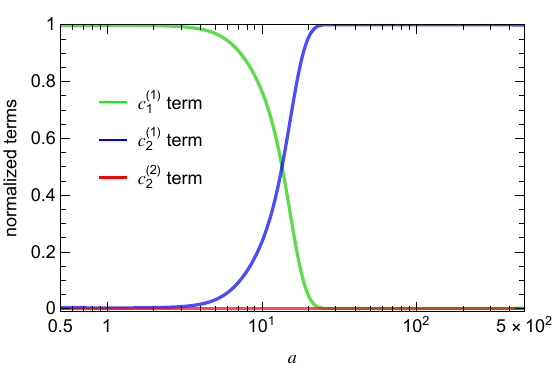}}\hfill
\caption{Training results for the production example. (a): The ground truth value of the field amplitude, the prediction from the GlueNN, and the curve obtained from the $C^1$ matching procedure. The shaded band indicates the variation obtained by shifting the matching point by $\pm 20\%$. GlueNN closely reproduces the true solution, whereas the $C^{1}$ matching shows a clear deviation. (b): The normalized terms of the asymptotic solutions, which approach constant values toward each end of the domain. Here, each term in Eq.~\eqref{eq:sol_inf_prod} is normalized by $y(a)$ itself.}
\label{fig:grav_res}
\end{figure}

For training, we used a head network with five hidden layers of widths 1, 4, 4, 4, 50  and an output layer of width 50, and three trunk networks with one hidden layer of width 50 and an output layer of width 1. Also, we use the small argument approximation of the special functions, $H_{\nu}^{(1)}(x)\propto x^{-\nu}$, and $J_{\nu}(x)\propto x^{\nu}$, and therefore parameterize the output as
\begin{equation}
    y(a) = e^{c_1^{(1)}(a)}\frac{a}{a_\ast} +e^{c_2^{(1)}(a)}+e^{c_2^{(2)}(a)}\frac{ a_\ast}{a}
    \label{eq:sol_inf_prod}
\end{equation}
where $a_\ast =0.10$; $c_1^{(1)}$, $c_2^{(1)}$, and $c_2^{(2)}$ are the outputs of the three trunk networks. As in the previous example, we exponentiate the outputs to enforce positivity and to make it easier for the network to learn coefficients that span several orders of magnitude. Then the MSE losses are
\begin{align}
\mathrm{MSE}_{\mathrm{data}} &= \frac{\lambda_a}{N_a}\sum_{a\in\{a_\alpha\}}|y(a) - h(a)|^2,\\
    \mathrm{MSE}_{\mathcal{D}} &= \frac{\lambda_b}{N_b}\sum_{a\in\{a_\beta\}} \left|\mathcal{D}[y(a)]\right|^2,\\
\mathrm{MSE}_{\mathrm{patch}} &= \frac{\lambda_c}{N_c}\sum_{a\in\{a_\gamma\}} \left|e^{c_1^{(1)}(a)}\frac{a}{a_\ast}\right|^2,
\end{align}
where $\mathcal{D}$ is defined in Eq.~\eqref{eq:vectorEq} with  $H=150$, $k =2.0$, $m=0.10$.\footnote{For illustration, we rescale the parameters $H$, $k$, and $m$, which have units of mass, to dimensionless quantities.}

\begin{table}[tbp]
\centering
\caption{Sampling Intervals and Hyperparameters for Quantum Tunneling Experiment}
\label{tab:tunneling_params}
\setlength{\tabcolsep}{6pt} 
\begin{tabular}{ >{\centering\arraybackslash}p{0.10\columnwidth} | >{\centering\arraybackslash}p{0.56\columnwidth} | >{\centering\arraybackslash}p{0.18\columnwidth} }
\toprule
\textbf{Set}  & \textbf{Interval} & \textbf{Points ($N$)} \\
\midrule \midrule 
\multicolumn{3}{c}{\textit{Real Part Training}} \\ \midrule \midrule
$\{x_\alpha\}$ & $[6.05, 13.14]$ & $140$ \\
$\{x_\beta\}$ &  $[-13.14, 11.62]$ & $490$ \\ 
$\{x_\gamma\}$ &  $[-4.08, 4.03]$ & $160$ \\
$\{x_\delta\}$ &  $[-13.14, -5.09] \cup [4.03, 13.14]$ & $340$ \\ \midrule \midrule
\multicolumn{3}{c}
{\textit{Imaginary Part Training}} \\ \midrule\midrule
$\{x_\alpha\}$ & $[6.05, 13.14]$ & $140$ \\
$\{x_\beta\}$ & $[-13.14, 11.62]$ & $490$ \\ 
$\{x_\gamma\}$ & $[-3.06, 5.04]$ & $160$ \\
$\{x_\delta\}$ &  $[-13.14, -5.09] \cup [5.04, 13.14]$ & $320$ \\
\bottomrule 
\multicolumn{3}{p{0.95\columnwidth}}{\scriptsize Note: The loss weights were set to $\lambda_a=1.0$, $\lambda_b=3.0$, and $\lambda_c=\lambda_d=0.25$. All the sample sets are collected linearly in the given interval.}
\end{tabular}
\end{table}

We present the hyperparameter for the experiment in Table~\ref{tab:inflation_params}, and we present the training results in Fig.~\ref{fig:grav_res}. This example is intrinsically more challenging than the chemical-reaction case. The latter is governed by a first-order equation, thus the asymptotic behavior in each regime is given by a single analytic solution, so one does not need to choose between multiple independent solutions. By contrast, the present mode equation is second order, and there are two linearly independent solutions. Constructing the correct global solution therefore requires selecting the physically appropriate linear combination across the transition.

The GlueNN prediction for $X_L$ shows a successful agreement with the ground truth data (Fig.~\ref{fig:grav_res_a}). For comparison, we also show the curve obtained from the $C^{1}$ matching scheme, in which the two asymptotic solutions for $X_L$ are matched at the boundary $a = k/m$ by imposing the continuity of $X_L$ and its derivative $dX_L/da$; this $C^{1}$ matched solution exhibits noticeable deviations from the actual solution. Although the agreement can sometimes be improved by moving the matching point by hand, this is essentially guesswork, and there is no systematic or robust prescription for selecting an optimal matching location. GlueNN instead learns the transition directly from the governing equation, thereby providing a reliable and systematic alternative to manual matching even in this second-order case.

Figure~\ref{fig:grav_res_b} shows the relative contribution of each term in the asymptotic solution in Eq.~\eqref{eq:X_L_asymt} as a function of $a$. As anticipated, the asymptotic solution valid for $a \ll k/m$ dominates at small $a$, a smooth transition occurs between the two regimes, and the term proportional to $c^{(1)}_{2}$ dominates for $a \gg k/m $. These results further confirm that GlueNN correctly capture the expected transitional behavior of the solution.

\subsection{Quantum Tunneling}

In our last example, we consider a standard quantum–mechanical
tunneling problem, which have broad applications ranging from the scanning tunneling microscope~\cite{BINNIG1983236}, electron transport in semiconductor devices~\cite{RANUAREZ20061939}, Josephson junctions~\cite{PhysRevLett.47.265,PhysRevLett.55.1908} and first-order phase transition in field theory~\cite{Devoto_2022}.

As a simple illustration, we study the one–dimensional Schrödinger equation for an incident plane wave on a localized potential barrier,
\begin{equation}{\label{eq:scrhodingerEq}}
\mathcal{D}[\psi(x)]=-\frac{\hbar^2}{2m} \frac{d^2\psi}{dx^2}+ V(x)\psi - E\psi=0.
\end{equation}
This setup provides a simple but nontrivial testbed, since the wave function reduces to plane waves far away from the barrier, while in the vicinity of the barrier it is governed by a different asymptotic form. By choosing a boater-shaped potential
\begin{equation}
V(x) = V_0 \left( \frac{1}{e^{-(x+d/2)/\sigma}+1} + \frac{1}{e^{(x-d/2)/\sigma} +1} -1\right),
\end{equation}
where $d\gg \sigma$ and $E < V_0$, we obtain a barrier of
height $V_0$ and width $d$ with smooth edges. For $|x|\gg d/2$, the equation reduces to the free particle equation, and its solution is a superposition of plane waves,
\begin{equation}
\psi(x)= Ae^{ikx} + B e^{-ikx},
\end{equation}
where $k^2=2mE/\hbar^2$. On the other hand, for $|x| \ll d/2 $, the potential is approximately constant, $V(x)\simeq V_0$,  so the asymptotic solution becomes
\begin{equation}
 \psi(x) = C  e^{\kappa x} +D  e^{-\kappa x},
\end{equation}
where $\kappa^2 = 2m (V_0-E)/\hbar^2 $. Here, $A, B, C, D\in\mathbb{C}$. In order to determine the reflection and transmission coefficients, one can impose an arbitrary initial condition on the transmission side of the barrier. Due to the linearity of the Schrödinger equation, physical observables such as a reflection and transmission coefficient depend solely on amplitude ratios, making them invariant under the specific amplitude and phase of the outgoing wave, i.e., specific initial condition.

For training, we used a network with a head network with three hidden layers of widths 100, 100, 100, and an output layer of width 100, and four trunk networks with one hidden layer of width 100 and an output layer of width 1. For numerical convenience, the wave functions decomposed into its real and imaginary components, which can be solved independently. Therefore, parameterization of the outputs becomes
\begin{equation}
    \begin{split}
        &\mathrm{Re}[\psi(x)]=c_1^{(1)}\cos(kx) + c_1^{(2)} \sin(kx)+c_2^{(1)}  e^{\kappa x} +c_2^{(2)}  e^{-\kappa x},\\
        &\mathrm{Im}[\psi(x)]=c_3^{(1)}\cos(kx) + c_3^{(2)} \sin(kx)+c_4^{(1)}  e^{\kappa x} +c_4^{(2)}  e^{-\kappa x},
    \end{split}
    \label{eq:sol_tunneling}
\end{equation}
where all coefficients are real.
For the real part, the MSE losses are constructed as
\begin{align}
\mathrm{MSE}_{\mathrm{data}} &= \frac{\lambda_a}{N_a}\sum_{x\in\{x_\alpha\}}|\mathrm{Re}[\psi(x)] - h(a)|^2,\\
    \mathrm{MSE}_{\mathcal{D}} &= \frac{\lambda_b}{N_b}\sum_{x\in\{x_\beta\}} \left|\frac{\mathcal{D}[\mathrm{Re}[\psi(x)]]}{|\mathrm{Re}[\psi(x)]|+\epsilon}\right|^2,\\
\mathrm{MSE}_{\mathrm{patch}}^{(1)} &= \frac{\lambda_c}{N_c}\sum_{x\in\{x_\gamma\}} \left|c_1^{(1)}\cos(kx) + c_1^{(2)} \sin(kx)\right|^2,\\
\mathrm{MSE}_{\mathrm{patch}}^{(2)} &= \frac{\lambda_d}{N_d}\sum_{x\in\{x_\delta\}} \left|c_2^{(1)}  e^{\kappa x} +c_2^{(2)}  e^{-\kappa x}\right|^2,
\end{align}
where $\mathcal{D}$ is defined in Eq.~\eqref{eq:scrhodingerEq} with $\sigma=0.5$, $d=10$, $V_0=4.1$, $k =2.0$, $m=0.5$, and to avoid division by zero, we add $\epsilon = 1.0$ to the denominator of the differential loss term.\footnote{For illustration, we rescale the parameters $V_0$, $k$, and $m$, which have units of mass, to dimensionless quantities, and let $\hbar = 1$.} The loss for the imaginary part of the outputs can be obtained by replacing
$\{c_1^{(1)},\, c_1^{(2)},\,c_2^{(1)},\,c_2^{(2)},\,\mathrm{Re[\psi]}\}$ in MSE losses into $\{c_3^{(1)},\,c_3^{(2)},\,c_4^{(1)},\,c_4^{(2)},\,\mathrm{Im}[ \psi ]\}$, respectively.

\begin{figure*}
\centering
\subfloat[\label{fig:1d_q_ex_res_a}]{\includegraphics[width=0.48\linewidth]{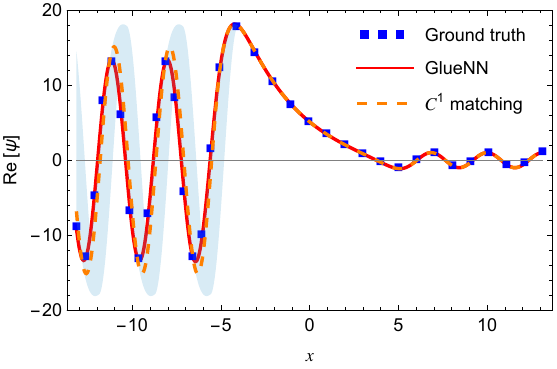}}\hfill
\subfloat[\label{fig:1d_q_ex_res_b}]{\includegraphics[width=0.48\linewidth]{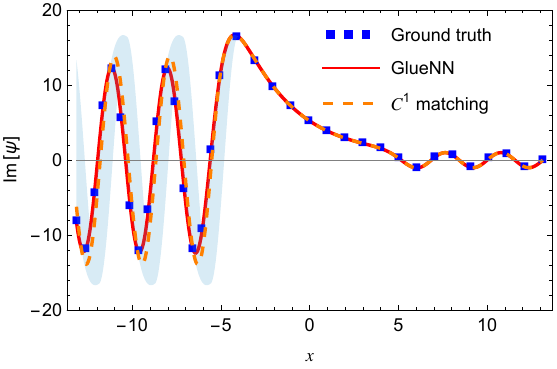}}\hfill
\subfloat[\label{fig:1d_q_ex_res_c}]{\includegraphics[width=0.48\linewidth]{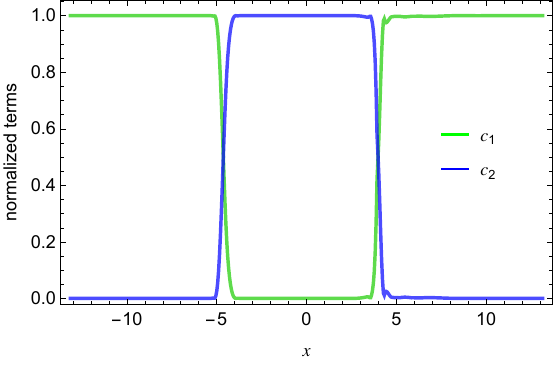}}\hfill
\subfloat[\label{fig:1d_q_ex_res_d}]{\includegraphics[width=0.48\linewidth]{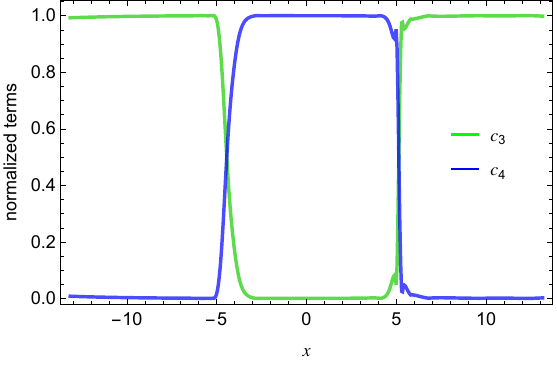}}\hfill
\caption{Training results for the 1-D quantum tunneling. (a), (b): The ground truth value, the prediction from the GlueNN, and the curve obtained from the $C^1$ matching procedure for the real part and imaginary part of wave function, respectively. The shaded band indicates the variation obtained by shifting the matching point by $\pm 20\%$. GlueNN closely reproduces the true solution, whereas $C^{1}$ matching shows a clear deviation. (c), (d): The normalized terms of the asymptotic solutions for the real part and imaginary part, respectively, which approach constant values toward each end of the domain. Here, $c_1,\,c_2,\,c_3,\,c_4$ denote the normalized quantities defined by $c_{1,3}\equiv\sqrt{c_{1,3}^{(1)2}+c_{1,3}^{(2)2}}/n_{r,i}$ and $c_{2,4}\equiv(|c_{2,4}^{(1)}|e^{\kappa x}+|c_{2,4}^{(2)}|e^{-\kappa x})/n_{r,i}$, where $n_{r,i}\equiv\sqrt{c_{1,3}^{(1)2}+c_{1,3}^{(2)2}}+|c_{2,4}^{(1)}|e^{\kappa x}+|c_{2,4}^{(2)}|e^{-\kappa x}$. }
\label{fig:tun_res}
\end{figure*}

We present the hyperparameter for the experiment in Table~\ref{tab:tunneling_params}, and the training results in Fig.~\ref{fig:1d_q_ex_res_a}, Fig.~\ref{fig:1d_q_ex_res_b}. Unlike the previous examples, this setting is substantially more challenging, as it features three distinct regimes and multiple competing solution branches. Therefore, GlueNN must simultaneously identify the appropriate linear combination of two independent modes and learn the transitions across several regime boundaries within a single unified model.

The GlueNN prediction for $\mathrm{Re}[\psi]$ and $\mathrm{Im}[\psi]$  shows close agreement with the ground truth data, giving reflection and transmission probability as
\begin{equation}
    |R_\text{GlueNN}|^2=0.9911, \quad |T_\text{GlueNN}|^2=0.0121,
\end{equation}
compared to the ground truth values of $|R_\text{true}|^2=0.9876$ and $|T_\text{true}|^2=0.0124$. Consequently, GlueNN predicts the reflection and transmission coefficients with relative errors of $0.35 \%$ and $-2.4 \%$, respectively.
For comparison, we also show the curve obtained from the $C^{1}$ matching scheme, in which the two asymptotic solutions for $\psi$ are matched at the boundary $x = -d/2$ by imposing the continuity of $\psi$ and its derivative $d\psi/dx$; this $C^{1}$ matched solution exhibits noticeable deviations from the actual solution, giving reflective and transmission probability $|R|^2=0.9909_{-0.0006}^{+0.0064}$ and $|T|^2=0.0095_{-0.0029}^{+0.0040}$, respectively.
Here, central values of $|R|^2$ and $|T|^2$ are obtained by $C^{1}$ matching at the boundary $x=-d/2$, and asymmetric uncertainties are obtained by varying the matching point within a $\pm20\%$ range around the reference point. This uncertainty reflects a systematic dependence on the matching prescription rather than a statistical error.

Fig.~\ref{fig:1d_q_ex_res_c} and Fig.~\ref{fig:1d_q_ex_res_d} show the relative contribution of each term in the asymptotic solution in Eq.~\eqref{eq:sol_tunneling} as a function of $x$. As anticipated, the  oscillatory solution valid for $|x| \gg d/2$ dominates at large $|x|$, a smooth transition occurs between the two regimes, and the terms proportional to the exponential dominate for $|x| \ll d/2 $. Therefore, the training results from GlueNN correctly capture the expected behavior of the solution.

\section{Discussions}

In this paper, we propose the GlueNN framework, which uses deep neural networks to capture the evolution of generic solutions across different patches of a domain. Whereas standard PINNs are trained to return the value of the solution field at each point in the domain, GlueNN instead targets the underlying analytic structure by directly learning the undetermined coefficients of asymptotic solutions of the differential equation. This coefficient-based approach is particularly useful because, in many applications, these coefficients are the physically relevant quantities that may encode conserved properties or possess a direct physical interpretation.

We demonstrate the applicability of this framework in three representative settings: chemical reactions, inflationary particle production, and quantum tunneling. In the chemical-reaction example, we consider a hot thermal plasma transitioning from equilibrium to a non-equilibrium configuration. GlueNN accurately predicts the final yield of the target species and captures the crossover between the relevant asymptotic regimes.
 Because the yield is exponentially sensitive to the point at which the system departs from equilibrium, a naive $C^0$ matching procedure is generally unable to achieve comparable robustness, whereas GlueNN provides a stable and systematic alternative. 
In the cosmological example, we consider the production of a massive vector boson during inflation, where the mode equation is a second order differential equation. This example highlights GlueNN’s ability to handle differential equations with multiple linearly independent solutions, where a particular linear combination must be selected. GlueNN reproduces the ground-truth solution and improves upon a naive $C^1$ matching.
In the quantum tunneling example, we apply the framework to a quantum mechanical problem governed by the Schrödinger equation, where the wavefunction exhibits oscillatory behavior in the far field and exponential decay within the potential barrier. This setting features multiple regime crossings, requiring the solution to transition between qualitatively different behaviors more than once across the domain. The results demonstrate that GlueNN can learn these multi-regime transitions in a single unified model and accurately reconstruct the physical wavefunction.

We also observe several practical aspects of the framework. First, the relative weighting between each MSE term in the loss must be chosen carefully to balance data against enforcing the governing equation. Each term plays a distinct role: the data loss imposes the initial and/or boundary conditions, the differential equation loss enforces the global dynamics, and the patchwise suppression term supports the transition between patch-specific asymptotic solutions. Note that the patch-wise suppression term is not always necessary. 
Second, a proper normalization of each loss term may be necessary. Because the losses are averaged over many sample points while the solution and its derivatives can vary by many orders of magnitude across the domain, unnormalized terms can be dominated by a limited subset of points. An appropriate normalization helps weight different regions more uniformly and thereby better captures the global behavior. 

More broadly, the GlueNN paradigm is not limited to the specific examples considered here. Many problems in physics, engineering, and applied mathematics admit families of asymptotic or basis solutions whose coefficients encode the essential physical content of the system. In such problems, recasting the learning task in terms of these coefficients rather than the full field can provide a compact, interpretable representation. We therefore expect that GlueNN could be applied to broader fields across science and engineering.

\vspace{2mm}
\section*{acknowledgments}
This work was supported in part by the National Research Foundation of Korea (Grant No. RS-2024-00352537). JY acknowledges support from the KAIST Jang Young Sil Fellow Program.

\bibliographystyle{IEEEtran}
\bibliography{ref.bib}

@article{RAISSI2019686,
title = {Physics-informed neural networks: A deep learning framework for solving forward and inverse problems involving nonlinear partial differential equations},
journal = {Journal of Computational Physics},
volume = {378},
pages = {686-707},
year = {2019},
issn = {0021-9991},
doi = {https://doi.org/10.1016/j.jcp.2018.10.045},
url = {https://www.sciencedirect.com/science/article/pii/S0021999118307125},
author = {M. Raissi and P. Perdikaris and G.E. Karniadakis}
}

@article{routray2025enforcingasymptoticbehaviordnns,
      title={Enforcing asymptotic behavior with DNNs for approximation and regression in finance}, 
      author={Hardik Routray and Bernhard Hientzsch},
      journal={arXiv:2411.05257},
      url={https://arxiv.org/abs/2411.05257}
}

@article{Antonov2020AsymptoticsControl,
  author  = {Antonov, Alexandre and Konikov, Michael and Piterbarg, Vladimir},
  title   = {Neural Networks with Asymptotics Control},
  journal = {SSRN},
  year    = {2020},
  url     ={https://ssrn.com/abstract=3544698}
}

@article{MICHOSKI2020193,
title = {Solving differential equations using deep neural networks},
journal = {Neurocomputing},
volume = {399},
pages = {193-212},
year = {2020},
issn = {0925-2312},
doi = {https://doi.org/10.1016/j.neucom.2020.02.015},
url = {https://www.sciencedirect.com/science/article/pii/S0925231220301909},
author = {Craig Michoski and Miloš Milosavljević and Todd Oliver and David R. Hatch}
}

@article{Jeong:2025omu,
    author = "Jeong, Hyun-Sik and Kim, Hanse and Kim, Keun-Young and Yun, Gaya and Yu, Hyeonwoo and Yun, Kwan",
    title = "{AdS/Deep-Learning made easy II: neural network-based approaches to holography and inverse problems}",
    journal = "arXiv:2511.22522",
    reportNumber = "APCTP Pre2025 - 024",
    month = "11",
    year = "2025"
}

@article{Song:2020agw,
    author = "Song, Mugeon and Oh, Maverick S. H. and Ahn, Yongjun and Kima, Keun-Young",
    title = "{AdS/Deep-Learning made easy: simple examples}",
    eprint = "2011.13726",
    archivePrefix = "arXiv",
    primaryClass = "physics.class-ph",
    doi = "10.1088/1674-1137/abfc36",
    journal = "Chin. Phys. C",
    volume = "45",
    number = "7",
    pages = "073111",
    year = "2021"
}

@article{jin2022asymptoticpreservingneuralnetworksmultiscale,
      title={Asymptotic-Preserving Neural Networks for Multiscale Time-Dependent Linear Transport Equations}, 
      author={Shi Jin and Zheng Ma and Keke Wu},
      journal={arXiv:2111.02541},
      url={https://arxiv.org/abs/2111.02541}
}

@article{Bertaglia2022Asymptotics,
author = {Bertaglia, Giulia and Lu, Chuan and Pareschi, Lorenzo and Zhu, Xueyu},
title = {Asymptotic-Preserving Neural Networks for multiscale hyperbolic models of epidemic spread},
journal = {Mathematical Models and Methods in Applied Sciences},
volume = {32},
number = {10},
pages = {1949-1985},
year = {2022},
doi = {10.1142/S0218202522500452},
URL = {
        https://doi.org/10.1142/S0218202522500452
}
}

@article{Amirhossein2023Theory,
title = {Theory-guided physics-informed neural networks for boundary layer problems with singular perturbation},
journal = {Journal of Computational Physics},
volume = {473},
pages = {111768},
year = {2023},
issn = {0021-9991},
doi = {https://doi.org/10.1016/j.jcp.2022.111768},
url = {https://www.sciencedirect.com/science/article/pii/S0021999122008312},
author = {Amirhossein Arzani and Kevin W. Cassel and Roshan M. D'Souza}
}

@article{Pratama2023EXPLORING,
    author = {Muchamad Pratama and Agus Gunawan},
    title = {EXPLORING PHYSICS-INFORMED NEURAL NETWORKS FOR SOLVING BOUNDARY LAYER PROBLEMS},
    journal = {Journal of Fundamental Mathematics and Applications (JFMA)},
year={2023},
  volume = {6},
    number = {2},
   issn = {2621-6035},   
pages = {101--116}, 
doi = {10.14710/jfma.v6i2.20084},
    url = {https://ejournal2.undip.ac.id/index.php/jfma/article/view/20084}
}

@article{wang2024aspinnasymptoticstrategysolving,
      title={ASPINN: An asymptotic strategy for solving singularly perturbed differential equations}, 
      author={Sen Wang and Peizhi Zhao and Tao Song},
      journal={arXiv:2409.13185},
      url={https://arxiv.org/abs/2409.13185}, 
}

@article{JAGTAP2020113028,
title = {Conservative physics-informed neural networks on discrete domains for conservation laws: Applications to forward and inverse problems},
journal = {Computer Methods in Applied Mechanics and Engineering},
volume = {365},
pages = {113028},
year = {2020},
issn = {0045-7825},
doi = {https://doi.org/10.1016/j.cma.2020.113028},
url = {https://www.sciencedirect.com/science/article/pii/S0045782520302127},
author = {Ameya D. Jagtap and Ehsan Kharazmi and George Em Karniadakis}
}

@article{osti_2282003,
  author       = {Jagtap, Ameya D. and Karniadakis, George Em},
  title        = {Extended Physics-Informed Neural Networks (XPINNs): A Generalized Space-Time Domain Decomposition Based Deep Learning Framework for Nonlinear Partial Differential Equations},
  doi          = {10.4208/cicp.oa-2020-0164},
  url          = {https://www.osti.gov/biblio/2282003},
  journal      = {Communications in Computational Physics},
  issn         = {ISSN 1815-2406},
  number       = {5},
  volume       = {28},
  place        = {United States},
  publisher    = {Global Science Press},
  year         = {2020},
  month        = {11}}

@article{Huang2024MultiScale,
author = {Huang, Jianlin and Qiu, Rundi and Wang, Jingzhu and Wang, Yiwei},
year = {2024},
month = {03},
pages = {100496},
title = {Multi-scale physics-informed neural networks for solving high Reynolds number boundary layer flows based on matched asymptotic expansions},
volume = {14},
journal = {Theoretical and Applied Mechanics Letters},
doi = {10.1016/j.taml.2024.100496}
}

@article{Graham:2015rva,
    author = "Graham, Peter W. and Mardon, Jeremy and Rajendran, Surjeet",
    title = "{Vector Dark Matter from Inflationary Fluctuations}",
    eprint = "1504.02102",
    archivePrefix = "arXiv",
    primaryClass = "hep-ph",
    doi = "10.1103/PhysRevD.93.103520",
    journal = "Phys. Rev. D",
    volume = "93",
    number = "10",
    pages = "103520",
    year = "2016"
}

@article{LIU2025113669,
title = {Asymptotic-preserving neural networks for the semiconductor Boltzmann equation and its application on inverse problems},
journal = {Journal of Computational Physics},
volume = {523},
pages = {113669},
year = {2025},
issn = {0021-9991},
doi = {https://doi.org/10.1016/j.jcp.2024.113669},
url = {https://www.sciencedirect.com/science/article/pii/S0021999124009173},
author = {Liu Liu and Yating Wang and Xueyu Zhu and Zhenyi Zhu},
}

@article{zhu2025deepasymptoticexpansionmethod,
      title={Deep asymptotic expansion method for solving singularly perturbed time-dependent reaction-advection-diffusion equations}, 
      author={Qiao Zhu and Dmitrii Chaikovskii and Bangti Jin and Ye Zhang},
      journal={arXiv:2505.23002},
      url={https://arxiv.org/abs/2505.23002}, 
}

@article{shen2025matchedasymptoticexpansionsbasedtransferable,
      title={Matched Asymptotic Expansions-Based Transferable Neural Networks for Singular Perturbation Problems}, 
      author={Zhequan Shen and Lili Ju and Liyong Zhu},
      journal={arXiv:2505.08368},
      url={https://arxiv.org/abs/2505.08368}
}

@article{sun2025pvdonetmultiscaleneuraloperator,
      title={PVD-ONet: A Multi-scale Neural Operator Method for Singularly Perturbed Boundary Layer Problems}, 
      author={Tiantian Sun and Jian Zu},
      journal={arXiv:2507.21437},
      url={https://arxiv.org/abs/2507.21437}
}

@article{Pradanya2025Parameter,
title = {A parameter-driven physics-informed neural network framework for solving two-parameter singular perturbation problems involving boundary layers},
journal = {Advances in Computational Science and Engineering},
volume = {5},
number = {0},
pages = {72-102},
year = {2025},
doi = {10.3934/acse.2025019},
url = {https://www.aimsciences.org/article/id/68d0f8140aae4966f7dc0712},
author = {Pradanya Boro and Aayushman Raina and Srinivasan Natesan}
}

@article{verma2025cosmologyinformedneuralnetworksinfer,
      title={Cosmology-informed Neural Networks to infer dark energy equation-of-state}, 
      author={Anshul Verma and Shashwat Sourav and Pavan K. Aluri and David F. Mota},
      journal={arXiv:2508.12032},
      url={https://arxiv.org/abs/2508.12032}, 
}

@article{hu2024betterneuralpdesolvers,
      title={Better Neural PDE Solvers Through Data-Free Mesh Movers}, 
      author={Peiyan Hu and Yue Wang and Zhi-Ming Ma},
      journal={arXiv:2312.05583},
      url={https://arxiv.org/abs/2312.05583}
}

@inproceedings{Huang2023NeuralStagger,
  title = 	 {{N}eural{S}tagger: Accelerating Physics-constrained Neural {PDE} Solver with Spatial-temporal Decomposition},
  author =       {Huang, Xinquan and Shi, Wenlei and Meng, Qi and Wang, Yue and Gao, Xiaotian and Zhang, Jia and Liu, Tie-Yan},
  booktitle = 	 {Proceedings of the 40th International Conference on Machine Learning},
  pages = 	 {13993--14006},
  year = 	 {2023},
  editor = 	 {Krause, Andreas and Brunskill, Emma and Cho, Kyunghyun and Engelhardt, Barbara and Sabato, Sivan and Scarlett, Jonathan},
  volume = 	 {202},
  series = 	 {Proceedings of Machine Learning Research},
  month = 	 {Jul},
  publisher =    {PMLR}
}

@article{Bunch:1978yq,
    author = "Bunch, T. S. and Davies, P. C. W.",
    title = "{Quantum Field Theory in de Sitter Space: Renormalization by Point Splitting}",
    doi = "10.1098/rspa.1978.0060",
    journal = "Proc. Roy. Soc. Lond. A",
    volume = "360",
    pages = "117--134",
    year = "1978"
}

@inproceedings{
bafghi2023pinnstorch,
title={{PINN}s-Torch: Enhancing Speed and Usability of Physics-Informed Neural Networks with PyTorch},
author={Reza Akbarian Bafghi and Maziar Raissi},
booktitle={The Symbiosis of Deep Learning and Differential Equations III},
year={2023},
url={https://openreview.net/forum?id=nl1ZzdHpab}
}

@article{BINNIG1983236,
title = {Scanning tunneling microscopy},
journal = {Surface Science},
volume = {126},
number = {1},
pages = {236-244},
year = {1983},
issn = {0039-6028},
doi = {https://doi.org/10.1016/0039-6028(83)90716-1},
url = {https://www.sciencedirect.com/science/article/pii/0039602883907161},
author = {G. Binnig and H. Rohrer}
}

@article{RANUAREZ20061939,
title = {A review of gate tunneling current in MOS devices},
journal = {Microelectronics Reliability},
volume = {46},
number = {12},
pages = {1939-1956},
year = {2006},
issn = {0026-2714},
doi = {https://doi.org/10.1016/j.microrel.2005.12.006},
url = {https://www.sciencedirect.com/science/article/pii/S0026271406000205},
author = {Juan C. Ranuárez and M.J. Deen and Chih-Hung Chen}
}

@article{PhysRevLett.47.265,
  title = {Macroscopic Quantum Tunneling in 1-\ensuremath{\mu}m Nb Josephson Junctions},
  author = {Voss, Richard F. and Webb, Richard A.},
  journal = {Phys. Rev. Lett.},
  volume = {47},
  issue = {4},
  pages = {265--268},
  numpages = {0},
  year = {1981},
  month = {Jul},
  publisher = {American Physical Society},
  doi = {10.1103/PhysRevLett.47.265},
  url = {https://link.aps.org/doi/10.1103/PhysRevLett.47.265}
}

@article{PhysRevLett.55.1908,
  title = {Measurements of Macroscopic Quantum Tunneling out of the Zero-Voltage State of a Current-Biased Josephson Junction},
  author = {Devoret, Michel H. and Martinis, John M. and Clarke, John},
  journal = {Phys. Rev. Lett.},
  volume = {55},
  issue = {18},
  pages = {1908--1911},
  numpages = {0},
  year = {1985},
  month = {Oct},
  publisher = {American Physical Society},
  doi = {10.1103/PhysRevLett.55.1908},
  url = {https://link.aps.org/doi/10.1103/PhysRevLett.55.1908}
}

@article{Devoto_2022,
doi = {10.1088/1361-6471/ac7f24},
url = {https://doi.org/10.1088/1361-6471/ac7f24},
year = {2022},
month = {aug},
publisher = {IOP Publishing},
volume = {49},
number = {10},
pages = {103001},
author = {Devoto, Federica and Devoto, Simone and Di Luzio, Luca and Ridolfi, Giovanni},
title = {False vacuum decay: an introductory review},
journal = {Journal of Physics G: Nuclear and Particle Physics}
}

@article{HU2023107183,
title = {Augmented Physics-Informed Neural Networks (APINNs): A gating network-based soft domain decomposition methodology},
journal = {Engineering Applications of Artificial Intelligence},
volume = {126},
pages = {107183},
year = {2023},
issn = {0952-1976},
doi = {https://doi.org/10.1016/j.engappai.2023.107183},
url = {https://www.sciencedirect.com/science/article/pii/S0952197623013672},
author = {Zheyuan Hu and Ameya D. Jagtap and George Em Karniadakis and Kenji Kawaguchi}
}

@book{Kolb:1990vq,
    author = "Kolb, Edward W. and Turner, Michael S.",
    title = "{The Early Universe}",
    reportNumber = "FERMILAB-BOOK-1990-01",
    doi = "10.1201/9780429492860",
    isbn = "978-0-429-49286-0, 978-0-201-62674-2",
    publisher = "Taylor and Francis",
    volume = "69",
    month = "5",
    year = "2019"
}

@article{Holdom:1985ag,
    author = "Holdom, Bob",
    title = "{Two U(1)'s and Epsilon Charge Shifts}",
    reportNumber = "UTPT-85-30",
    doi = "10.1016/0370-2693(86)91377-8",
    journal = "Phys. Lett. B",
    volume = "166",
    pages = "196--198",
    year = "1986"
}

@article{Fabbrichesi:2020wbt,
    author = "Fabbrichesi, Marco and Gabrielli, Emidio and Lanfranchi, Gaia",
    title = "{The Dark Photon}",
    eprint = "2005.01515",
    archivePrefix = "arXiv",
    primaryClass = "hep-ph",
    doi = "10.1007/978-3-030-62519-1",
    month = "5",
    year = "2020"
}

@article{Caputo:2021eaa,
    author = "Caputo, Andrea and Millar, Alexander J. and O'Hare, Ciaran A. J. and Vitagliano, Edoardo",
    title = "{Dark photon limits: A handbook}",
    eprint = "2105.04565",
    archivePrefix = "arXiv",
    primaryClass = "hep-ph",
    reportNumber = "NORDITA-2021-036",
    doi = "10.1103/PhysRevD.104.095029",
    journal = "Phys. Rev. D",
    volume = "104",
    number = "9",
    pages = "095029",
    year = "2021"
}

\clearpage
\vfill

\end{document}